\documentclass[10pt,twocolumn,letterpaper]{article}

\usepackage[pagenumbers]{wacv} 

\usepackage{graphicx}
\usepackage{amsmath}
\usepackage{amssymb}
\usepackage{booktabs}
\usepackage{multirow}
\usepackage[dvipsnames]{xcolor}
\usepackage{pifont}
\newcommand{\cmark}{\ding{51}}%
\newcommand{\xmark}{\ding{55}}%
\usepackage{array}

\usepackage[pagebackref,breaklinks,colorlinks]{hyperref}

\let\titleold\title
\renewcommand{\title}[1]{\titleold{#1}\newcommand{\thetitle}{#1}}
\def\maketitlesupplementary
   {
   \newpage
       \twocolumn[
        \centering
        \Large
        \textbf{\thetitle}\\
        \vspace{0.5em}Supplementary Material \\
        \vspace{1.0em}
       ] 
   }

\usepackage[capitalize]{cleveref}
\crefname{section}{Sec.}{Secs.}
\Crefname{section}{Section}{Sections}
\Crefname{table}{Table}{Tables}
\crefname{table}{Tab.}{Tabs.}

\def\confName{WACV}
\def\confYear{2024}

\begin{document}

\title{BoostRad: Enhancing Object Detection by Boosting Radar Reflections}

\author{Yuval Haitman\footnotemark[1] \space  and Oded Bialer\thanks{Both authors contributed equally to this work.\\ Both authors are with General Motors, Yuval Haitman is also with the School of Electrical and Computer Engineering in Ben Gurion University of the Negev.} \\
General Motors, 
Technical Center Israel\\
{\tt\small haitman@post.bgu.ac.il, \tt\small oded.bialer8@gmail.com}}

\maketitle

\begin{abstract}
Automotive radars have an important role in autonomous driving systems. The main challenge in automotive radar detection is the radar's wide point spread function (PSF) in the angular domain that causes blurriness and clutter in the radar image. Numerous studies suggest employing an \textit{'end-to-end'} learning strategy using a Deep Neural Network (DNN) to directly detect objects from radar images. This approach implicitly addresses the PSF's impact on objects of interest. In this paper, we propose an alternative approach, which we term ''Boosting Radar Reflections'' (\textit{BoostRad}). In \textit{BoostRad}, a first DNN is trained to narrow the PSF for all the reflection points in the scene. The output of the first DNN is a boosted reflection image with higher resolution and reduced clutter, resulting in a sharper and cleaner image. Subsequently, a second DNN is employed to detect objects within the boosted reflection image. We develop a novel method for training the boosting DNN that incorporates domain knowledge of radar's PSF characteristics. \textit{BoostRad's} performance is evaluated using the RADDet and CARRADA datasets, revealing its superiority over reference methods.
\end{abstract}
\vspace{-15pt}
\section{Introduction}
\label{sec:intro}
Automotive radars play an important role in autonomous driving systems, offering extended detection range and resilience to adverse weather and lighting conditions. These radars emit RF signals, which interact with objects and bounce back to the receiving antennas as echoes. By applying signal processing algorithms \cite{bilik2016automotive}, the received signals are coherently combined over a brief duration (e.g., 20ms) and across multiple antennas to generate a radar image. This image, characterized by dimensions of range and angle, captures the intensity of the radar's received energy at range and angle coordinates of each pixel. Serving as a visual representation of the scene, the radar image is further processed using computer vision algorithms to extract valuable scene information, thereby enhancing the perception of the environment.

\begin{figure}[t]
  \centering
   \includegraphics[width=1\linewidth]{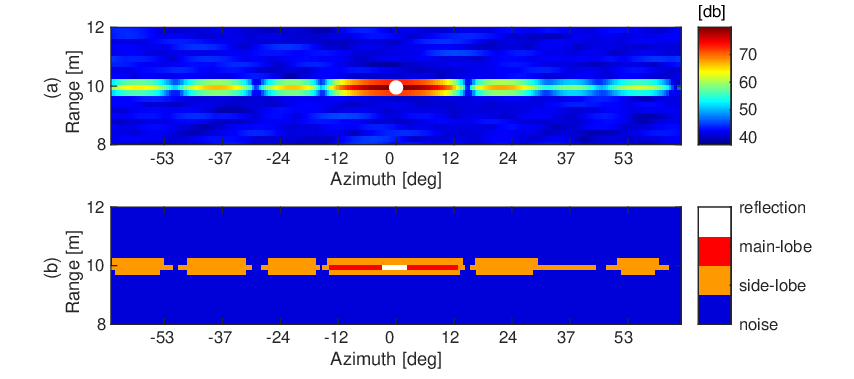}
   \vspace{-20pt}
   \caption{Radar's point spread function (PSF).}
   \label{fig:radar_spread_func}
   \vspace{-5pt}
\end{figure}
\begin{figure}[t]
  \centering
   \includegraphics[width=1\linewidth]{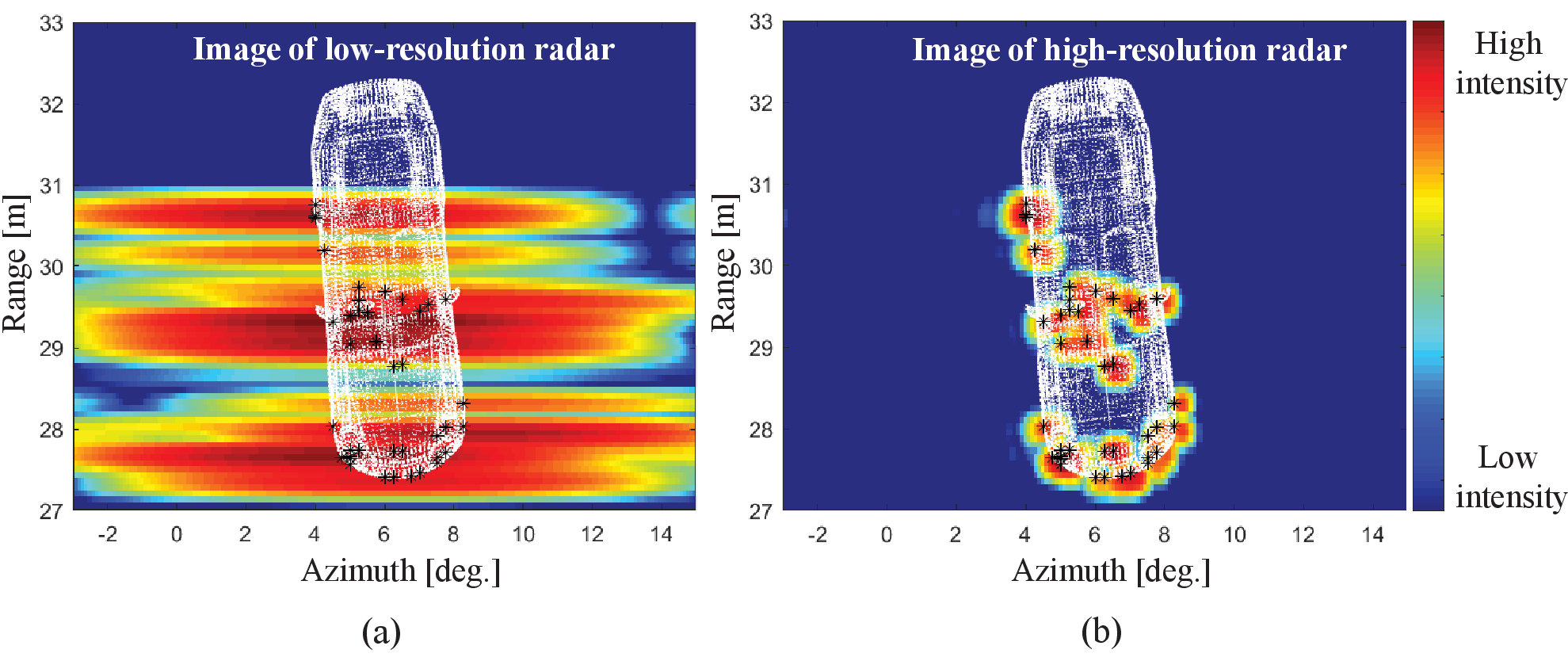}
    \vspace{-20pt}
   \caption{Comparison of low-resolution realistic automotive radar reflection image (a) and high-resolution 'super-radar' image (b). White points depict vehicle, black asterisk shows radar reflection point from vehicle. Reflection points in (a) have wide PSF, while in (b) the PSF is narrow.}
   \label{fig:high_re_low_res_ex}
   \vspace{-8pt}
\end{figure}

One of the key obstacles in object detection from radar images is the presence of blurriness and excessive clutter noise. These issues arise due to the wide point spread function (PSF) of the radar in the angular domain, which is a consequence of its coarse angular resolution. 
Fig.\ref{fig:radar_spread_func}(a) illustrates the spreading function of a single reflection point, represented by a white point. The PSF consists of a main-lobe centered on the reflection point, along with multiple angular side-lobes that are attenuated replicas of the main-lobe at angle offsets from the main-lobe. These main-lobe and side-lobe components are depicted with different colors in Fig.\ref{fig:radar_spread_func}(b). In radar images, reflections from objects appear as multiple points, each exhibiting a wide PSF in the angular domain. Fig.~\ref{fig:high_re_low_res_ex}(a) shows a radar image of a simulated car (white dots) with black asterisks marking the reflection points. The PSF centered at each black asterisk has an angular resolution of $8^{\circ}$, which is commonly found in production-grade automotive radars. The wide main-lobe of the PSF causes image blurriness, making it difficult to accurately determine object shape, orientation and class, as well as discriminating between closely located objects. Additionally, the PSF's side lobes contribute to image clutter in scenarios involving multiple objects, resulting in both miss-detection of objects with low reflection intensity and false detection of clutter as actual objects. 

Fig.\ref{fig:high_re_low_res_ex}(b) illustrates the radar image for the same reflection points as in Fig.\ref{fig:high_re_low_res_ex}(a). However, in this case, the radar used has an angular resolution of $0.1^{\circ}$, resulting in a PSF with significantly narrower main-lobe and lower side-lobes. It is clear that this narrower PSF allows for better estimation of object characteristics and eliminates clutter. Nevertheless, achieving such high angular resolution through hardware in automotive radars presents practical challenges. It would necessitate a larger physical size and a greater number of antennas, making it difficult to mount on vehicles while also increasing system complexity and cost \cite{bialer2021super}. Therefore, it is crucial to explore algorithmic alternatives that can reduce the PSF of practical radars, such as demonstrated in Fig.~\ref{fig:high_re_low_res_ex}(a).

Several notable works \cite{kim2020yolo,redmon2016you,dong2020probabilistic,zhang2021raddet,bochkovskiy2020yolov4,zhang2020object,meyer2021graph,stroescu2020combined,wang2021rodnet} have proposed an \textit{'end-to-end'} learning approach for object detection, which involves training a Deep Neural Network (DNN) to estimate objects directly from the radar image using ground truth reference bounding boxes. In this approach, the adverse effects of the radar's PSF on the objects of interest are implicitly mitigated. In contrast, our paper presents a distinct approach called Boosting Radar Reflections (\textit{BoostRad}). In \textit{BoostRad}, a boosting DNN is trained to narrow the main-lobe and reduce the side-lobes of the PSF for all reflection points in the scene. The output, referred to as the boosted image, is a higher-resolution and cleaner radar image with reduced clutter. Subsequently, a second DNN is employed to perform object detection from  the boosted image. The boosting DNN aims to transform a low-resolution image, illustrated in Fig.~\ref{fig:high_re_low_res_ex}(a), into a representation that closely resembles the high-resolution radar image depicted in Fig.\ref{fig:high_re_low_res_ex}(b). This transformation enables enhanced object detection capabilities. 

We present a novel approach for training the boosting DNN, distinguished by two key aspects. Firstly, it utilizes a deep learning technique that incorporates domain knowledge of the physical PSF characteristics of the radar. This is achieved through the integration of a unique ground truth reference of a high-resolution radar image with narrow PSF that is converted to a reflection probability map, and a customized loss function, enabling effective narrowing of the PSF. Secondly, to overcome the practical issue of acquiring real reference images with narrow PSF, we have developed a radar simulation that generates synthetic data. This allows us to train the boosting DNN exclusively on synthetic samples. Notably, the trained model demonstrates promising performance when applied to real radar images.

The performance of \textit{BoostRad} was assessed using the RADDet  \cite{zhang2021raddet} and CARRADA \cite{ouaknine2021carrada} datasets, which comprise automotive radar images taken from various scenarios. The evaluation clearly demonstrates the performance advantage of \textit{BoostRad} over multiple reference \textit{'end-to-end'} object detection methods.

The main contributions of this paper are:
\begin{enumerate}
    \item A novel technique that narrows the PSF in radar images, leading to improved object detection. This technique stands out due to its deep learning approach that incorporates domain knowledge of the physical sensor's PSF characteristics. The method can also be extended to tackle the wide PSF issue in other image-producing sensors, such as ultrasound, MRI, CT, telescopes, and low-end cameras. 
    
    \item Insight into the debate between \textit{'end-to-end'} and multi-stage object detection approaches. The paper challenges the prevailing trend of \textit{'end-to-end'} object detection approaches by demonstrating the superior performance of multi-stage methods in radar images. By enhancing the image prior to object detection, the proposed technique surpasses \textit{'end-to-end'} approaches, prompting further investigation of multi-stage approaches in the field of computer vision.

    \item Highlighting the value of synthetic simulation data for radar images.
    This paper successfully trained a DNN to narrow the PSF using synthetic data alone. The demonstrated effectiveness of the trained DNN on real radar data serves as a motivation for further exploration and utilization of radar simulation in computer vision. Additionally, the details of the developed radar simulation are disclosed to facilitate its utilization, thereby promoting further progress in radar-based computer vision research.      
\end{enumerate}

\section{Related Work} \label{sec:rel_work}

\subsection{Radar Object Detection}
One approach to integrate DNNs into radar object detection involves using radar detection points, known as the radar point cloud, as input for the DNN \cite{danzer20192d, schumann2018semantic, scheiner2019radar, feng2019point, kraus2020using, dreher2020radar}. These points are identified using the Constant False Alarm Rate (CFAR) algorithm \cite{rohling1983radar}, which detects peaks in the radar image surpassing a local noise threshold. However, CFAR introduces information loss during subsequent DNN processing.

To address the information loss caused by CFAR, an alternative approach directly applies DNNs to the radar image for object detection, bypassing CFAR. Kim et al. \cite{kim2020yolo} utilized YOLO \cite{redmon2016you} on radar images, demonstrating better performance than the conventional CFAR-based method. Xu et al. \cite{dong2020probabilistic} adapted a ResNet-18 encoder with a CNN decoder to output 3D bounding box center, size, and orientation. RADDet \cite{zhang2021raddet} employed residual blocks and YOLO detection heads \cite{bochkovskiy2020yolov4}. Zhang et al. \cite{zhang2020object} used a CNN version of U-Net for radar object detection. Meyer et al. \cite{meyer2021graph} proposed graph convolution networks. Additional enhancements have been made by integrating temporal processing \cite{stroescu2020combined, wang2021rodnet} and by fusing radar and camera images \cite{hwang2022cramnet, kim2020low, bijelic2020seeing, lim2019radar, feng2020deep, nabati2021centerfusion, meyer2019deep, nabati2020radar}.

\subsection{Radar Datasets} 
Numerous publicly available automotive radar datasets with object detection annotations exist, which can be categorized to two groups based on the type of radar data they offer. The first group includes datasets with radar point cloud data (the CFAR detections), such as nuScenes \cite{caesar2020nuscenes}, Radar Scenes \cite{schumann2021radarscenes}, Pointillism \cite{bansal2020pointillism}, Cooperative Radars \cite{wwt7-w739-22}, aiMotive \cite{matuszka2022aimotivedataset}, and Astyx \cite{meyer2019automotive}.

The second group consists of datasets providing radar reflection intensity images before CFAR processing. The RADIATE dataset \cite{sheeny2021radiate} captures unique $360^{\circ}$ scanning radar reflection images, distinguishing it from conventional automotive radars with antenna arrays. The CRUW dataset \cite{wang2021rodnet2} offers radar images with a limited range of up to $25m$ and the annotations include object center points without bounding box information. RADIal \cite{rebut2022raw} and K-Radar \cite{paek2022k} contain higher-resolution radar reflection intensity images, but low level radar hardware details are undisclosed. As a result, simulating the radar and generating synthetic data for these datasets is unattainable.

The CARRADA dataset \cite{ouaknine2021carrada} provides radar reflection images from 30 controlled scenarios in an open environments with few objects per scene using a Texas Instruments (TI) automotive radar prototype \cite{TI_PROTOTYPE_1,TI_PROTOTYPE_2}. RADDet \cite{zhang2021raddet} offers TI radar-derived reflection intensity images across 15 diverse automotive scenarios. Both datasets have up to $50m$ range and 2D bounding box annotations. An important distinction lies in the availability of hardware specifications for the TI prototype radar used in these datasets. This availability enables radar simulation and the generation of synthetic radar data, offering a notable advantage.

\section{Object Detection With Reflection Boosting}\label{sec:method}
\begin{figure}[t]
  \centering
   \includegraphics[width=1\linewidth]{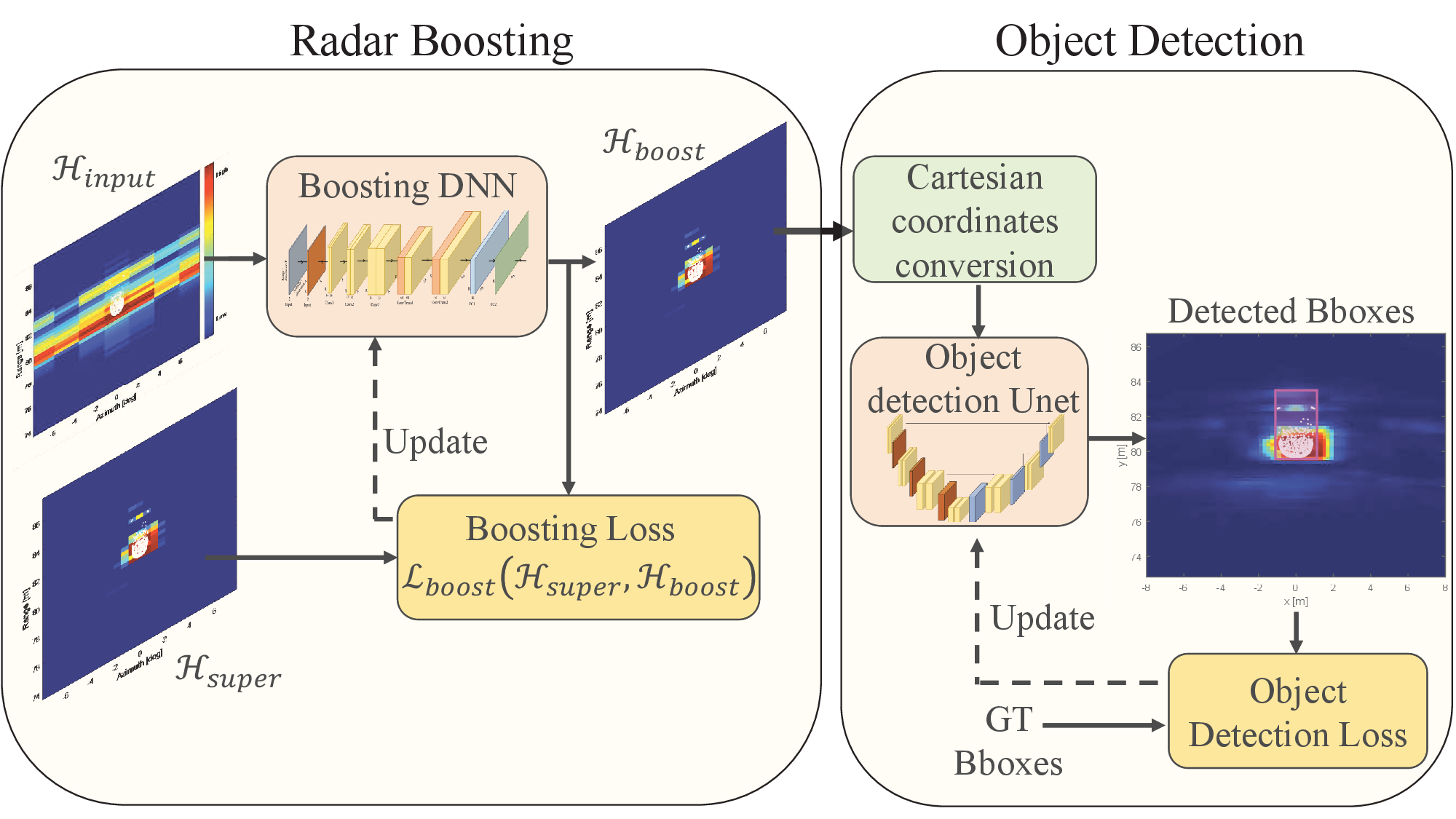}
   \caption{BoostRad Overview: The boosting DNN enhances input reflection image $\mathcal{H}_{input}$ to a radar image $\mathcal{H}_{boost}$ with a narrower PSF, resembling the 'super-radar' image $\mathcal{H}_{super}$. Subsequently, object detection DNN identifies objects in the boosted image.}
   \label{fig:full_dnn_scheme}
   \vspace{-5pt}
\end{figure}

A block diagram of the proposed object detection system is depicted in Fig.~\ref{fig:full_dnn_scheme}. The input radar image with range and angle dimensions undergoes processing by the reflection boosting DNN, which enhances resolution and reduces clutter by narrowing the main-lobe and damping side-lobes of the PSF across all  reflection points in the scene. Subsequently, another DNN detects objects within the boosting DNN output.

The boosting DNN is trained with a ground truth reference image from a high-resolution radar referred to as the 'super-radar'. This radar captures images from the same scene as the automotive radar but with a 
PSF featuring a narrower main-lobe and lower side-lobes.
However, a notable challenge lies in the unavailability of a 'super-radar' hardware. Moreover, the selection of 'super-radar' resolution enhancement factor and a suitable loss function play a crucial role in successfully training the boosting DNN to closely align automotive radar images with the 'super-radar' images. These challenges are addressed comprehensively in Sections \ref{sec:boosting_net} and \ref{sec:data_set} that follow.
 
\subsection{Reflection Boosting Network}\label{sec:boosting_net}
\begin{figure}[t]
  \centering
   \includegraphics[width=1\linewidth]
   {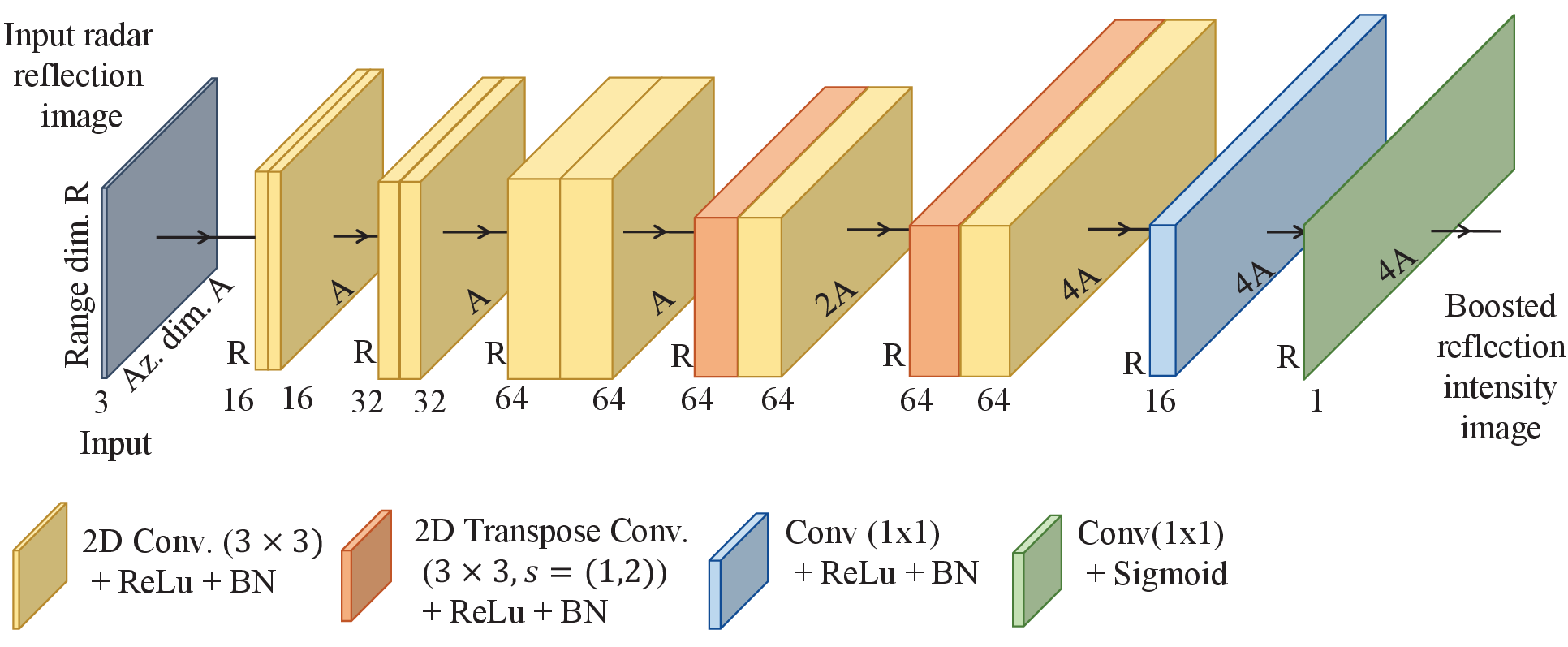} 
   \caption{Diagram depicting reflection boosting DNN architecture with 4x expansion in output angular dimension compared to input.}
   \label{fig:boost_dnn_scheme}
   \vspace{-10pt}
\end{figure}

\textbf{Architecture.} The reflection boosting network is depicted in Fig.~\ref{fig:boost_dnn_scheme}. The input to the network is the reflection intensity image with a uniform grid in the dimensions range and sinus of the azimuth angle. In this coordinate system, the radar's PSF is spatially invariant and separable in range and angle dimensions. These properties are essential for narrowing the PSF. While our method primarily addresses radars without elevation data, its applicability can extend to radars featuring azimuth and elevation by increasing the input tensor's dimensions. The input image comprises three channels: the real and imaginary parts of the highest intensity along the Doppler domain for each range-angle pixel, and the Doppler frequency corresponding to the most intense Doppler bin. In this paper, we refer to the latter channel as the Doppler map.

The boosting network is a 2D convolutional neural network tailored to preserve the radar image's range dimension while expanding the azimuth angular dimension. This approach retains the fine radar range resolution (usually below $25cm$) while increasing the initially lower angular resolution. This enhancement in angular resolution is achieved through successive layers of transpose convolutions in the angular domain. The network's output head consists of two layers of $1\times 1$ convolution filters followed by the Sigmoid function.
\newline \indent \textbf{Ground Truth Reference and Loss Function.} The boosting network's training involves generating the probability of reflection for each output range-azimuth pixel, with an azimuth resolution $\kappa$ times higher than the input. The ground truth reference is a reflection image constructed from a 'super-radar' that corresponds to the input image scenario.
The 'super-radar' image angular resolution is $\kappa$ times greater than the input radar image. As a result, the PSF in the 'super-radar' image is narrower by factor $\kappa$ and has lower side-lobes than the input radar image. 
However, obtaining a practical 'super-radar' is unfeasible due to the complex and costly requirements, including a larger aperture with numerous antennas. To circumvent this limitation, we've developed a radar simulation that generates synthetic pairs of automotive and 'super-radar' images originating from the same scene. These synthetic pairs are subsequently employed for training the boosting DNN. A comprehensive explanation of this simulation process is provided in Section \ref{sec:data_set}.
\newline \indent The proposed ground truth reference holds two significant advantages. Firstly, it shares the same sensor physical modality as the radar that produced the input image. In both images, the positions and intensity of reflection points remain consistent; the sole distinction lies in the spreading functions. Notably, this differs from using a LIDAR as a reference, where the reflection points of the LIDAR deviate considerably from the radar's. Secondly, this reference allows for the careful selection of achievable angular resolution, serving as a realistic goal for the network. While the ideal scenario would entail an extremely high $\kappa$, akin to LIDAR output resolution, physical constraints limit achievable output resolution, and complete elimination of the radar spreading function remains unattainable. As shown in the ablation study in Section \ref{sec:ablation_study}, selecting an excessively high $\kappa$ results in a decline in performance.
\newline \indent For calculating the loss function, we transform the 'super-radar' reflection intensity for each range and angle pixel into reflection probability. Let the intensity at the $n^{th}$ pixel of the 'super-radar' image be denoted as $z_n$.
We make the assumption that $z_n$ follows a chi-square distribution \cite{eaves2012principles} with variance $\sigma_s^2$ when the pixel contains a reflection point (where $\sigma_s^2$ varies with range), and has a noise variance of $\sigma_n^2$ when the pixel lacks a reflection point. Under this assumption, we deduce that the ground truth reference probability is formulated as
\vspace{-5pt}
\begin{equation}\label{eq:p0}
p_n=\frac{e^{-|z_n|^2/(2\sigma_s^2)}}{e^{-|z_n|^2/(2\sigma_s^2)}+\frac{\sigma_s^2}{\sigma_n^2}e^{-|z_n|^2/(2\sigma_n^2)}}.
\vspace{-5pt}
\end{equation}
For a more comprehensive explanation of the derivation of \eqref{eq:p0}, please refer to the supplementary material. 

The reflection boosting DNN is trained with a weighted cross-entropy loss between the pixel-wise probabilities of the boosting DNN output image and the ground truth reference image. The loss function is given by 
\begin{multline}\label{eq:loss_boost}
\mathcal{L}_{boost}=\rho_n\sum_{n\in \Omega_n} \mathcal{L}_{bce}(p_n,\hat{p}_n)+\\
\rho_r\sum_{n\in \Omega_r} \mathcal{L}_{bce}(p_n,\hat{p}_n)+\rho_s\sum_{n\in \Omega_s} \mathcal{L}_{bce}(p_n,\hat{p}_n),
\end{multline}
where $\mathcal{L}_{bce}(p_n,\hat{p}_n)$ is the cross entropy loss between $p_n$ and $\hat{p}_n$. The notations in \eqref{eq:loss_boost} are as follows. The symbol $n$ is the boosting DNN output pixel index, $\hat{p}_n$ is the DNN estimated probability of a reflection in the $n^{th}$ pixel, and $p_n$ is 
$n^{th}$ pixel reflection probability from the ground truth reference image given in \eqref{eq:p0}. 
The symbols $\Omega_n$, $\Omega_r$, and $\Omega_s$ denote three different sets of pixels, which will be explained below, and $\rho_n,\rho_r,\rho_s$ correspond to weight factors associated with each respective set.

The loss function in \eqref{eq:loss_boost} comprises three loss function terms corresponding to three sets of pixels from the input radar image denoted by $\Omega_n$, $\Omega_r$ and $\Omega_s$. The set $\Omega_r$ is the image's pixels containing reflection points. The set $\Omega_s$ are pixels belonging to a reflection point's spreading function but not the reflection point's pixel itself. These pixels can be distinguished as pixels not included in $\Omega_r$ that have  intensity above a noise level threshold. For example, we set the threshold 8 dB above the noise standard deviation (calculated empirically). The set $\Omega_n$ comprises pixels containing noise; these are the remaining pixels that do not belong to the sets $\Omega_r$ and $\Omega_s$. We refer the reader to Fig.~\ref{fig:radar_spread_func}(b) for an illustration of the pixel sets. The set $\Omega_r$ is colored white in the figure, the set $\Omega_s$ includes the red (main-lobe) and orange (side-lobes) pixels, and the set $\Omega_n$ is colored blue. 

The partitioning into these three sets facilitates the assignment of distinct weight factors ($\rho_r$, $\rho_s$, $\rho_n$) to the corresponding loss terms for each set. Notably, this partitioning empowers the enhancement of the weight factor associated with the pixels within the spreading function ($\rho_s$), thereby compelling the boosting DNN to effectively narrow the spreading function and reduce its side-lobes. The effectiveness of this enhancement will be illustrated in the ablation study detailed in Section \ref{sec:ablation_study}.

\subsection{Object Detection Network}\label{sec:obj_det_dnn}
The object detection DNN features four input channels. The first channel comprises the output of the reflection boosting DNN, converted to Cartesian coordinates. This transformation is important for object detection since object shape and size remain consistent in these coordinates. The second channel contains the Cartesian coordinates transformation of the Doppler map, which also serves as an input channel for the boosting DNN. Each pixel in the Doppler map represents the Doppler frequency of the highest-intensity Doppler bin within the Doppler domain of each range-angle pixel in the original reflection intensity image. The remaining two channels consist of the Cartesian coordinates $(x,y)$ for each cell. While object shapes remain invariant in Cartesian coordinates, their surface reflection intensity varies based on their position relative to the radar. Radar reflection intensity from surfaces depends on their angle and distance from the radar. This information relies on the object's position in relation to the radar. By providing Cartesian input coordinate values for each pixel, the network learns these relationships and applies them to object detection.

The object detection DNN is a standard U-Net architecture \cite{ronneberger2015u} that has an encoder-decoder network with skip connections. The encoder has ten convolution layers that gradually down-sample the spatial features by a factor of 16. The decoder has eight convolution layers that gradually up-sample the spatial features back to the original input dimension. Batch normalization and ReLU activation functions are applied after each layer. The final layer is a $1\times 1$ convolution layer, which outputs the probability of an object class (class score) and the 2D bounding box parameters of the vehicle in a bird's-eye view for each spatial cell. The bounding box specifications encompass the offset of the object center point from the cell center, the object's width and height, and the orientation represented by $cos(\theta), sin(\theta)$, with $\theta$ being the object's orientation angle. When the dataset lacks the bounding box's orientation angle, $\theta$ remains fixed at zero. During inference, non-maximal suppression is applied to filter out overlapping bounding boxes. The object detection network is trained using standard cross-entropy loss for classification and L2 regression loss for bounding box parameters. 

\subsection{Synthetic Data Generation}\label{sec:data_set}
Training the boosting DNN in Section \ref{sec:boosting_net} requires a unique ground truth reference reflection image of the 'super-radar'. However, the 'super-radar' hardware is not a commodity available and may not be practical for implementation due to the large aperture, high number of antennas, and system complexity. To overcome this limitation, we developed a high-fidelity radar simulation that generates synthetic radar reflection intensity images of realistic scenes for any desired radar specification.  

\begin{figure}[t]
  \centering
   \includegraphics[width=1\linewidth]{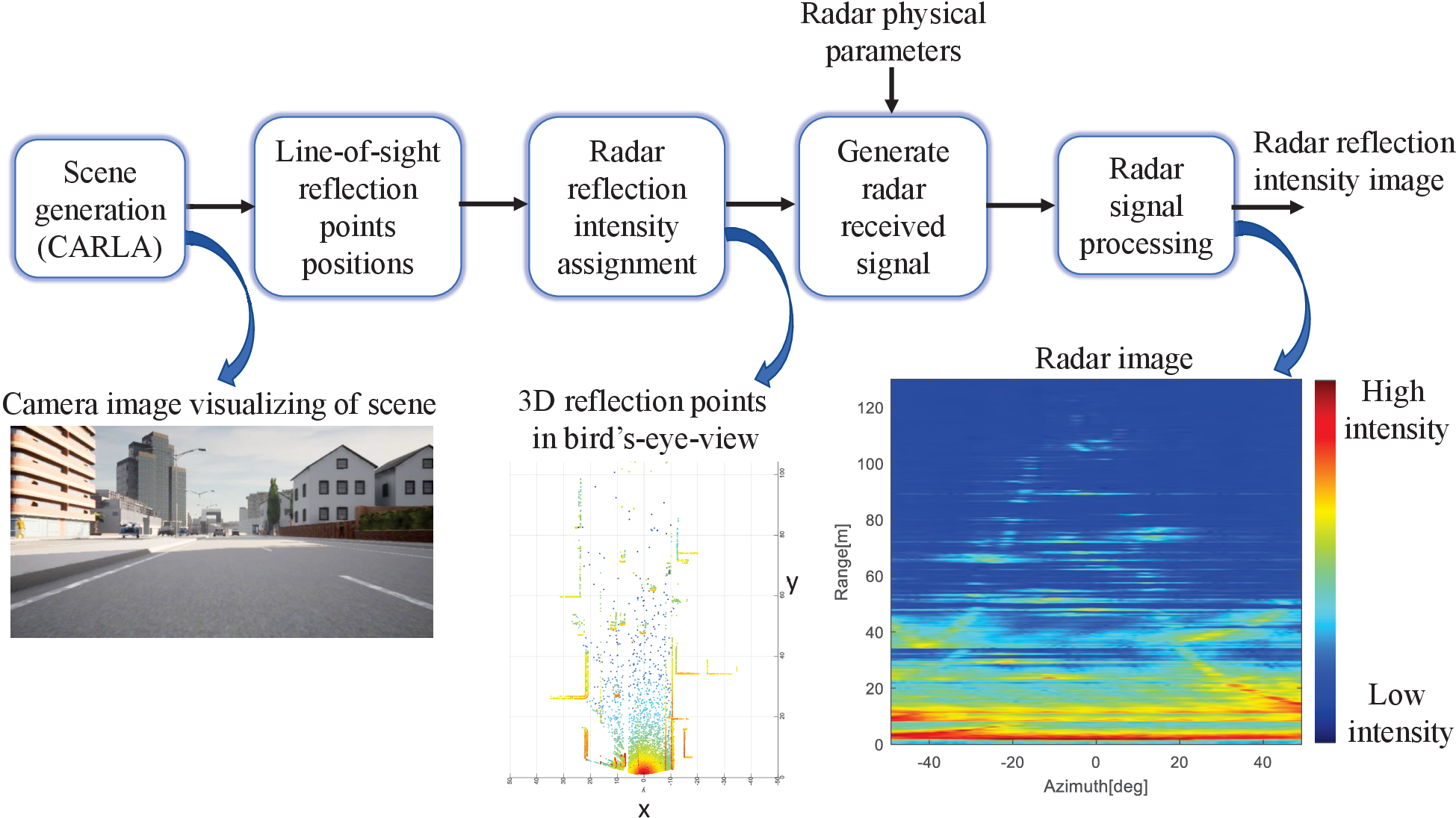}
   \caption{Radar simulation pipeline with illustrated simulated scenario: camera image, reflection points, and radar reflection image.}
   \label{fig:radar_processes}
   \vspace{-5pt}
\end{figure}

Fig.~\ref{fig:radar_processes} presents the simulation procedure for generating radar images, which is elaborated below. In the first part, the CARLA simulation \cite{Dosovitskiy17} generates a realistic driving scenario with roads, bridges, pavements, buildings, signs, poles, persons, vehicles, etc. From this CARLA simulation, a dense 3D point cloud is generated, capturing positions from all visible surfaces in the scene that can reflect radar signals. In the subsequent phase, each of these points is assigned a reflection intensity value using standard radar signal propagation and reflectivity formulas \cite{mahafza2003matlab}. The reflection point intensity is a function of three factors: (a) the surface material, (b) the angle between the normal vector of the point's surface and the direction from the point to the radar, and (c) the distance between the point and the radar. 

In the subsequent phase, the radar-received signal is synthesized using the reflection points and specific radar parameters, such as the antenna array layout and the transmitted waveform. The received signal is a sum of the received signals from all the reflection points in the scene with additional Gaussian noise. The received signal of each reflection point is generated by the point's reflection intensity, the distance between the point to the transmit and the receive antennas, and the radar's transmit waveform \cite{eaves2012principles, bilik2016automotive}. In the final step of the simulation, standard radar signal processing techniques are applied to the received signals collected from all antennas, resulting in the production of the radar reflection intensity image. For more details on the mathematical formulas utilized in the simulation's implementation, please refer to the supplementary material.

To enable boosting DNN inference on an actual physical radar, it's essential to train the DNN using synthetic simulation data that closely aligns with the physical radar characteristics. This simulation requires knowledge of the antenna array layout and the transmit waveform employed by the physical radar. For the ground truth reference, we simulate a higher resolution radar with narrow PSF, featuring an increased number of transmit and receive antennas spanning a broader aperture than the physical radar being tested.

\section{Experiments and Results}\label{sec:res}
To assess the performance of BoostRad, we utilize the RADDet \cite{zhang2021raddet} and CARRADA \cite{ouaknine2021carrada} datasets. Both of these datasets comprise radar reflection images alongside 2D bounding box annotations for objects. The radar images were captured using a TI prototype automotive radar \cite{TI_PROTOTYPE_1,TI_PROTOTYPE_2}, that has $50m$ range capability. The hardware specifications of the TI radar are publicly accessible, enabling us to simulate it and generate synthetic training data for the reflection boosting DNN (as detailed in Section \ref{sec:data_set}). These datasets provide comprehensive testing coverage for automotive radar scenarios. RADDet encompasses images from 15 different densely populated automotive scenarios under favorable weather conditions. On the other hand, CARRADA contains 30 staged scenarios with fewer objects per scenario, yet these are captured in varying conditions, including fair and more challenging weather conditions such as snow.

In the RADDet dataset utilized in \cite{zhang2021raddet}, the test and training images were sourced from the same scenarios, often with small time gaps between them. This split creates a strong correlation between test and training examples, leading to overfitting on the test set. To address this, we adjusted the RADDet data split for our performance evaluations, ensuring different scenarios in the train and test sets. Unlike RADDet, the CARRADA dataset didn't display this issue, so we retained the original data partitioning as proposed in \cite{ouaknine2021carrada}.
For dataset partitioning details, including ablation tests with the original RADDet split, see the supplementary material. 

\textbf{Implementation details.} 
The boosting and object detection models  were trained on single NVIDIA Tesla V100 GPU, for 100 epochs with batch size 32, and ADAM optimizer with learning rate of $10^{-4}$. More implementation details can be found in the supplementary material.

\subsection{Object Detection Performance Evaluation}\label{sec:res_object_det}
The boosting DNN's training set was generated by simulating the TI radar prototype used in RADDet and CARRADA datasets (see Section \ref{sec:data_set}). It included 15,000 synthetic radar images and their corresponding 'super-radar' reference images, with a 12 times higher angular resolution ($\kappa=12$) than the TI radar prototype. After training the boosting DNN, its weights were frozen for subsequent object detection training. Radar images from RADDet or CARRADA were passed through the boosting DNN, and its output replaced the original radar image for the object detection DNN. It's important to highlight that the same boosting DNN was utilized for both datasets. Conversely, the object detection DNN was optimized independently for RADDet and CARRADA, and their respective performances were evaluated separately.

We assess BoostRad's object detection performance against three reference methods: 'U-Net', 'RADDet', and 'Probabilistic'. The first method employs the same U-Net as BoostRad's, but with the original radar reflection image as input instead of the boosting DNN output image. 'RADDet' refers to the object detection network introduced in the RADDet paper \cite{zhang2021raddet}, while 'Probabilistic' refers to the network from \cite{dong2020probabilistic}. To examine the boosting DNN's generalization across different architectures, we also evaluated 'RADDet' and 'Probabilistic' using the boosting DNN output as input instead of the original radar reflection image

Table \ref{tab:results} summarizes the average precision (AP) results for the tested methods on RADDet and CARRADA at IOU thresholds 0.1 and 0.3. Supplementary material contains further results for larger IOU values. AP represents the area under the precision-recall curve and is provided for the 'car' and 'person' classes. In the car class, we also include trucks, which are larger vehicles such as vans. Cars are easier to detect than persons since their reflection intensity is significantly higher. To assess the performance on more challenging cases within the 'car' class, we also present the AP for cars at range (denoted by $R$) exceeding $40m$. 
The 'Boosting' in the second column specifies whether the input to the object detection network was the boosting DNN output (\cmark) or the original reflection image (\xmark). Green values indicate the AP difference between using boosting DNN output and original reflection image for the same detection method.

We first examine in Table \ref{tab:results} the results of BoostRad compared to the reference methods 'U-Net', 'Probalistic', and 'RADDet' when the reference methods use the original reflection intensity image and not the boosting DNN output. 
This comparison reveals that BoostRad consistently achieves higher AP values than the reference methods, across all IOU values, object classes, detection distances, and datasets. For class car, the performance advantage of BoostRad over the best reference method, 'U-Net', is relatively small when calculating the AP for all ranges. The reason is that most cars are at relatively close distances and could be relatively easily detected by all methods. However, in more challenging scenarios such as 'person' detection at all distances and 'car' detection beyond $40 m$ with an IOU of 0.3, BoostRad demonstrates a notable performance gain over the reference methods. 
'U-Net' stands out as the best-performing reference method, potentially due to its supplementary channels containing $(x,y)$ Cartesian coordinates, which contributes to object detection. Notably, the 'RADDet' method exhibits lower AP than the values reported in \cite{zhang2021raddet} due to differences in the train-test set split,  preventing overfitting (details in the supplementary material).

Next, we evaluate the performance gain using the boosting DNN output instead of the original reflection intensity image as input for the 'RADDet' and 'Probalistic' reference methods. The results in Table \ref{tab:results} show that the AP of both reference methods improves when using the boosting DNN output instead of the original reflection intensify image. The performance gain increases as the detection cases are harder, i.e., persons at all distances and cars at a distance greater than $40m$.

\begin{table*}[h]
\centering
\caption{Object Detection Average Precision on RADDet and CARRADA Datasets}
\vspace{-5pt}
\label{tab:results}
\begin{tabular}{>{\centering\arraybackslash}p{1cm}>{\centering\arraybackslash}p{0.67cm}p{0.83cm}p{0.83cm}p{0.83cm}p{0.83cm}p{0.83cm}p{0.83cm}p{0.83cm}p{0.83cm}p{0.83cm}p{0.83cm}p{0.83cm}p{0.83cm}}
\toprule
\multicolumn{2}{c}{} & \multicolumn{6}{c}{RADDet Dataset} & \multicolumn{6}{c}{CARRADA Dataset} \\
\cmidrule(lr){1-8} \cmidrule(lr){9-14}
\multicolumn{1}{c}{\multirow{2}{*}{}} &    & \multicolumn{4}{c}{Car}                                                         & \multicolumn{2}{c}{Person} & \multicolumn{4}{c}{Car}   & \multicolumn{2}{c}{Person}          \\
\cmidrule(lr){3-6} \cmidrule(lr){7-8} \cmidrule(lr){9-12} \cmidrule(lr){13-14}
\multicolumn{1}{c}{}                        &       & \multicolumn{2}{c}{$R \ge 0$} & \multicolumn{2}{c}{$R \ge 40$} & \multicolumn{2}{c}{$R \ge 0$} & \multicolumn{2}{c}{$R \ge 0$} & \multicolumn{2}{c}{$R \ge 40$} & \multicolumn{2}{c}{$R \ge 0$} \\
\cmidrule(lr){3-4} \cmidrule(lr){5-6} \cmidrule(lr){7-8} \cmidrule(lr){9-10} \cmidrule(lr){11-12} \cmidrule(lr){13-14}
\multicolumn{1}{c}{\multirow{1}{*}{Method}}                 & Boost. & @ 0.1       & @ 0.3             & @ 0.1              & @ 0.3             & @ 0.1              & @ 0.3 & @ 0.1       & @ 0.3             & @ 0.1              & @ 0.3             & @ 0.1              & @ 0.3            \\
\cmidrule(lr){1-8} \cmidrule(lr){9-14}
\multirow{3}{*}{RADDet} &  \multirow{1}{*}{\xmark} & 83.69 & 72.96 &  42.45 & 21.27 & 29.76 &  15.38 & 87.59 & 70.90 &  76.46 & 68.74 & 29.10 & 10.24\\
& \multirow{1}{*}{\cmark} & 85.91 & 78.19 &  51.65 & 28.25 & 31.50 & 18.90 & 88.48 & 73.79 &  80.10 & 78.02 & 35.59 & 22.39\\
& & \noindent\color{Green}{+2.22} & \noindent\color{Green}{+5.23} & \noindent\color{Green}{+9.20} & \noindent\color{Green}{+6.98} & \noindent\color{Green}{+1.74} & \noindent\color{Green}{+3.52} & \noindent\color{Green}{+0.89} & \noindent\color{Green}{+2.89} &  \noindent\color{Green}{+3.64} & \noindent\color{Green}{+9.28} & \noindent\color{Green}{+6.49} & \noindent\color{Green}{+12.15}\\
\cmidrule(lr){1-8} \cmidrule(lr){9-14}
\multirow{3}{*}{Probalistic} & \multirow{1}{*}{\xmark} & 83.32 & 74.80 &   43.28  &  28.50 & 21.30 & 17.10 & 84.72 & 67.56 &  69.40 & 63.92 & 22.23 & 10.99\\
& \multirow{1}{*}{\cmark} & 89.17 & 79.74 &  68.19 & 45.85 & 41.07 & 31.23 & 86.12 & 78.19 &  81.55 & 73.29 & 32.29 & 22.30\\
& & \noindent\color{Green}{+5.85} & \noindent\color{Green}{+4.94} &  \noindent\color{Green}{+24.91} & \noindent\color{Green}{+17.35} & \noindent\color{Green}{+19.77} & \noindent\color{Green}{+14.13} & \noindent\color{Green}{+1.40} & \noindent\color{Green}{+10.63} &  \noindent\color{Green}{+12.15} & \noindent\color{Green}{+9.37} & \noindent\color{Green}{+10.06} & \noindent\color{Green}{+11.31}\\
\cmidrule(lr){1-8} \cmidrule(lr){9-14}
\multirow{1}{*}{U-Net} & \multirow{1}{*}{\xmark} & 88.90 &  81.36 &   71.90 &   47.90  &  25.50 & 19.60  & 87.86 & 81.05 & 75.56 & 64.35 & 25.50 & 23.74\\
\cmidrule(lr){1-8} \cmidrule(lr){9-14}
\multirow{2}{*}{BoostRad} &  \multirow{2}{*}{\cmark} & \textbf{90.00} & \textbf{83.55} & \textbf{79.50} &  \textbf{62.10}     & \textbf{41.20} & \textbf{33.00}   & \textbf{88.86} & \textbf{81.43} &  \textbf{82.23} & \textbf{75.87} & \textbf{40.45} & \textbf{35.46}\\
&  & \noindent\color{Green}{+1.10} & \noindent\color{Green}{+2.19} &  \noindent\color{Green}{+7.60}  & \noindent\color{Green}{+14.20}  & \noindent\color{Green}{+15.70}  & \noindent\color{Green}{+13.40} & \noindent\color{Green}{+1.00} & \noindent\color{Green}{+0.38} &  \noindent\color{Green}{+6.67} & \noindent\color{Green}{+11.52} & \noindent\color{Green}{+14.95} & \noindent\color{Green}{+11.72}\\
\bottomrule
\end{tabular}
\end{table*}

\begin{figure*}[]
\vspace{-10pt}
  \centering
   \includegraphics[width=0.825\linewidth]{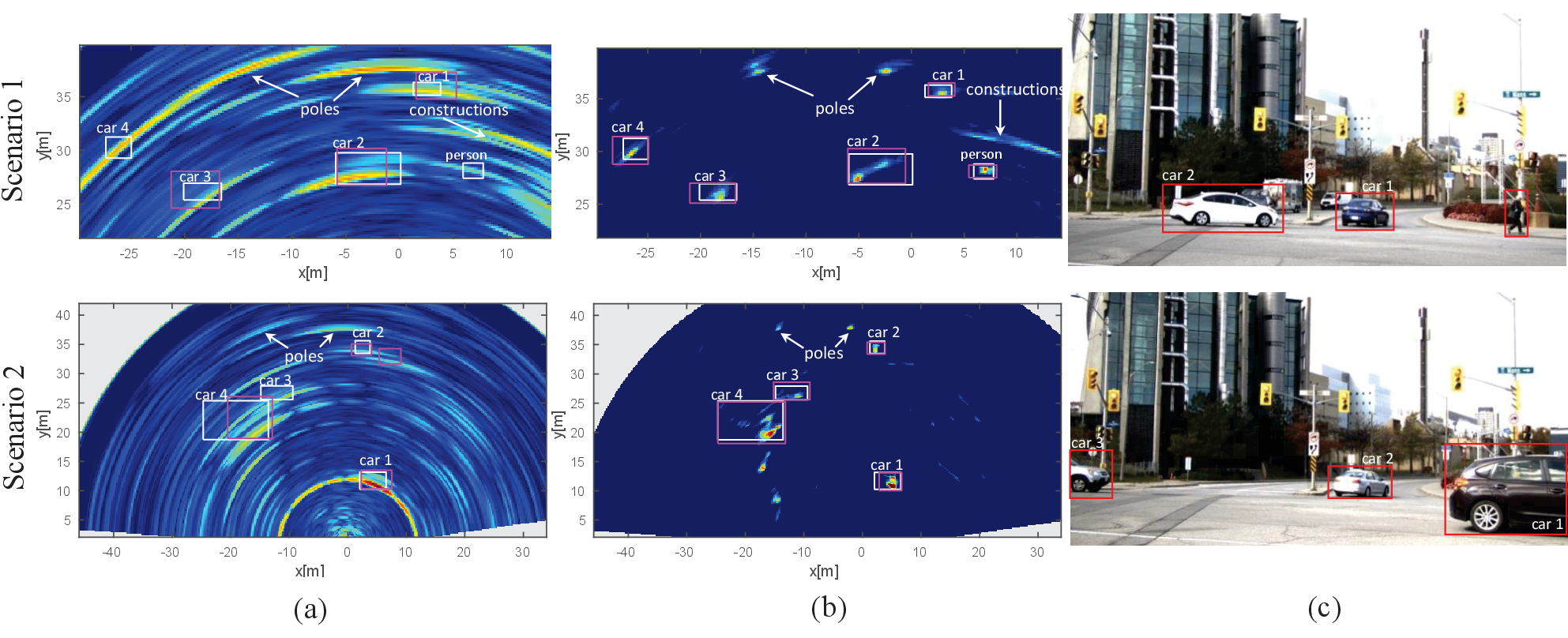}
   \vspace{-10pt}
   \caption{Qualitative examples from RADDet dataset. Each row is a different scenario. (a) original radar reflection image with the 'U-Net' reference detections, (b) boosting DNN output image with BoostRad detections, (c) camera image of the scene with ground truth bounding boxes. Detections in (a) and (b) are purple, while ground truth bounding boxes are white.}
   \label{fig:qualive_eval}
\end{figure*}

\begin{table}
\vspace{-10pt}
\centering
\caption{Boosting Ablation Study (RADDet, AP @0.3)}
\vspace{-10pt}
\label{tab:loss_ablation}
\begin{tabular}{cccccc}
\toprule
 & \multirow{2}{0.6cm}{GT Map.}  &        & \multicolumn{2}{c}{Car}                                                         & \multicolumn{1}{c}{Person}            \\
\cmidrule(lr){4-5} \cmidrule(lr){6-6}
\multirow{1}{0.6cm}{Loss} &  & \multirow{1}{0.6cm}{$\Omega_s$}      & \multicolumn{1}{c}{$R \ge 0$} & \multicolumn{1}{c}{$R \ge 40$} & \multicolumn{1}{c}{$R \ge 0$} \\
\midrule
\multirow{2}{0.6cm}{L1} & \multirow{1}{0.6cm}{\cmark} &  \multirow{1}{0.6cm}{ -} &  79.46 &  43.26  & 23.43 \\ 
 & \multirow{1}{0.6cm}{\cmark}  &  \multirow{1}{0.6cm}{\cmark} &  78.20  &  43.98 &  26.55 \\ 
\midrule
\multirow{3}{0.6cm}{CE} & \multirow{1}{0.6cm}{\cmark} &  \multirow{1}{0.6cm}{\xmark} &  81.54  &  50.64 &  25.98 \\
 & \multirow{1}{0.6cm}{\xmark} &  \multirow{1}{0.6cm}{\cmark} &   68.14  & 19.43 &   5.18 \\
 & \multirow{1}{0.6cm}{\cmark} & \multirow{1}{0.6cm}{\cmark} &   \textbf{83.55} &     \textbf{62.10} &  \textbf{33.00} \\
\midrule
\vspace{-15pt}
\end{tabular}
\end{table}

\begin{figure}[h]
\vspace{-15pt}
  \centering
   \includegraphics[width=1\linewidth]{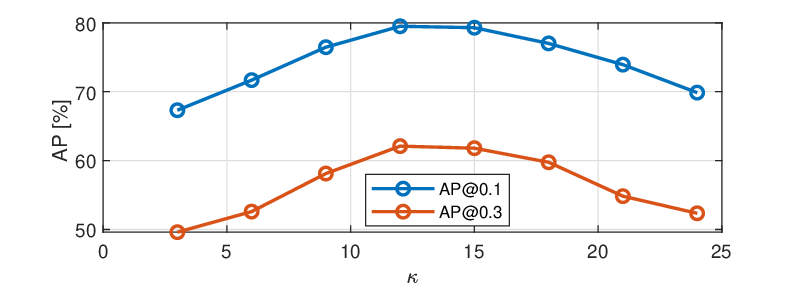}
      \vspace{-20pt}
   \caption{Effect of $\kappa$ on object detection. Raddet cars at $R>40$}
   \label{fig:kappa_fig}
   \vspace{-15pt}
\end{figure}

\subsection{Qualitative Performance Evaluation}\label{sec:qualitative_perf}
Fig.~\ref{fig:qualive_eval} provides a qualitative comparison between BoostRad and the 'U-Net' reference method, using two examples from the RADDet test set. Additional qualitative examples can be found in the supplementary material. Each row within the figure corresponds to a distinct scenario. For each scenario, we present the original radar reflection image (a), the image after passing through the boosting DNN (b), and the camera image of the scene (c). Detected bounding boxes from the 'U-Net' reference method are plotted in purple in (a), while those from BoostRad are plotted in purple in (b). Ground truth bounding boxes are indicated in white within (a) and (b), and in red in (c). Car and person bounding boxes are labeled with text and identification numbers. Other objects in the environment, such as poles and structures, are marked on the figure to aid orientation, even though they aren't part of the dataset and thus aren't classified.

From Fig.~\ref{fig:qualive_eval}, it's evident that the original radar reflection intensity image (a) appears blurred and cluttered, attributed to the radar's wide PSF. In contrast, the image at the boosting DNN's output (b) is sharper, cleaner, and displays a considerably narrower PSF. In the first scenario, BoostRad detects all objects, whereas 'U-Net' misses 'car 4' and the 'person'. This discrepancy might be due to differences between the boosting DNN image (b) and the reflection intensity image (a). Specifically, in scenario 1 (a), 'Car 4' is masked by the side-lobe of a pole at a similar range, and the person's reflection is relatively weak in comparison to nearby clutter arising from side-lobes of adjacent reflections. Moving to scenario 2, BoostRad successfully detects all objects, while 'U-Net' miss-detects 'car 3' and falsely detects a 'car' on the right side of 'car 2'. Such errors could stem from the higher side-lobes of the spreading function in (a) as opposed to (b). 

\subsection{Ablation Study}\label{sec:ablation_study}
The effective training of the boosting DNN can be attributed to its distinct loss function and ground truth reference, as examined in the ablation study outlined below. The boosting DNN loss function expressed in \eqref{eq:loss_boost} comprises two key components. The first component involves mapping the ground truth 'super-radar' reference intensity to probabilities using \eqref{eq:p0}, facilitating pixel-wise cross-entropy computation between it and the boosting DNN's output. The second component involves segmenting the pixel-wise cross-entropy into three sets of pixels ($\Omega_r,\Omega_s,\Omega_n$), each assigned distinct weights ($\rho_r,\rho_s,\rho_n$). This segmentation mechanism empowers the enforcement of PSF narrowing by amplifying the loss weight associated with the PSF's pixels.

Table \ref{tab:loss_ablation} presents a comparison of object detection average precision for IOU 0.3 on the RADDet dataset, evaluating the impact of training the boosting DNN with alternative loss functions from those employed in BoostRad. The table examines L1 loss instead of cross entropy, as indicated in the first column. Additionally, it investigates the use of cross entropy with binary mapping of the ground truth 'super-radar' reference intensity image via an optimized noise level threshold, instead of the probability mapping in \eqref{eq:p0}, as indicated in the second column. The table also compares results when partitioning cross-entropy pixels into sets $\Omega_r,\Omega_s,\Omega_n$ as in \eqref{eq:loss_boost}, separating side-lobe pixels from noise pixels, versus when partitioning only to signal and noise pixels, including side-lobe pixels in the noise category. The former case is indicated by \cmark \space and the later by \xmark \space  in the third column of the table, and a hyphen indicates the case of no segmentation to different pixel sets. 

The final row showcases performance with all proposed loss components of BoostRad. This configuration significantly outperforms alternative loss functions, underlining the significance of each component within the boosting DNN's loss function for achieving maximal object detection accuracy.

Next, we investigate the importance of selecting an appropriate resolution enhancement factor $\kappa$ for the 'super-radar' ground truth reference. Fig.~\ref{fig:kappa_fig} depicts the AP results for class 'car' at distance greater than $40m$ as a function of $\kappa$. The analysis reveals that $\kappa=12$ achieves optimal performance. A lower $\kappa$ fails to fully exploit the network's resolution enhancement capabilities. Conversely, selecting an excessively high $\kappa$ value leads to performance degradation, possibly due to attempting to surpass the physical limitations of resolution. Thus, the choice of $\kappa$ significantly affects performance, and its optimization was made possible through radar simulation, enabling the generation of ground truth references with various angular resolutions. 

Additional analysis of the boosting DNN image enhancement can be found in the supplementary material.

\section{Conclusion}\label{sec:conc}
We present a novel method to enhance radar image quality using a DNN that narrows the radar PSF, resulting in improved object detection. Our approach integrates domain knowledge, utilizing a high-resolution radar image with optimized resolution enhancement  ($\kappa$), a unique intensity-to-probability mapping, and a tailored cross entropy loss that enforces the attenuation of the PSF side-lobes. An ablation study confirms the importance of these components. Addressing the lack of narrow PSF radar hardware, we develop a simulation to generate synthetic data for the boosting DNN training. In testing on real data from RADDet and CARRADA datasets, \textit{BoostRad} outperforms reference methods in object detection.

Our work challenges prevailing \textit{'end-to-end'} object detection trends, encouraging exploration of multistage approaches. The PSF narrowing technique holds potential beyond radar, benefiting sensors with wide PSFs. Successful synthetic radar image utilization prompts further investigation of synthetic data for similar challenges in computer vision tasks involving radar and other sensors.

{\small
\bibliographystyle{ieee_fullname}
\bibliography{egbib}
}
\newpage
\appendix
\maketitlesupplementary
\section{Supplementary Performance Evaluation}
In this section we provide a performance analysis of \textit{BoostRad}, expanding upon the insights provided in Section \ref{sec:res}.
\subsection{Reflection Boosting Performance Evaluation}\label{sec:res_boost_dnn}
In this section, we assess the boosting DNN's accuracy in detecting reflection points. Notably, objects consist of multiple reflection points. Our evaluation involves comparing the boosting DNN's output to ground truth reflection points using a specific metric. The ground truth reflection points were determined as the central positions of pixels in the 'super-radar' reflection image with intensities surpassing a predefined noise threshold. On the other hand, detection points in the boosting DNN's output were considered as the centers of pixels surpassing a detection threshold. A precision-recall curve was generated for varying detection thresholds. Detection points within a 25cm radius of a ground truth point were considered true positive detections. The average precision (AP) was computed by the area under the precision-recall curve.

\begin{figure}[t]
  \centering
   \includegraphics[width=0.9\linewidth]{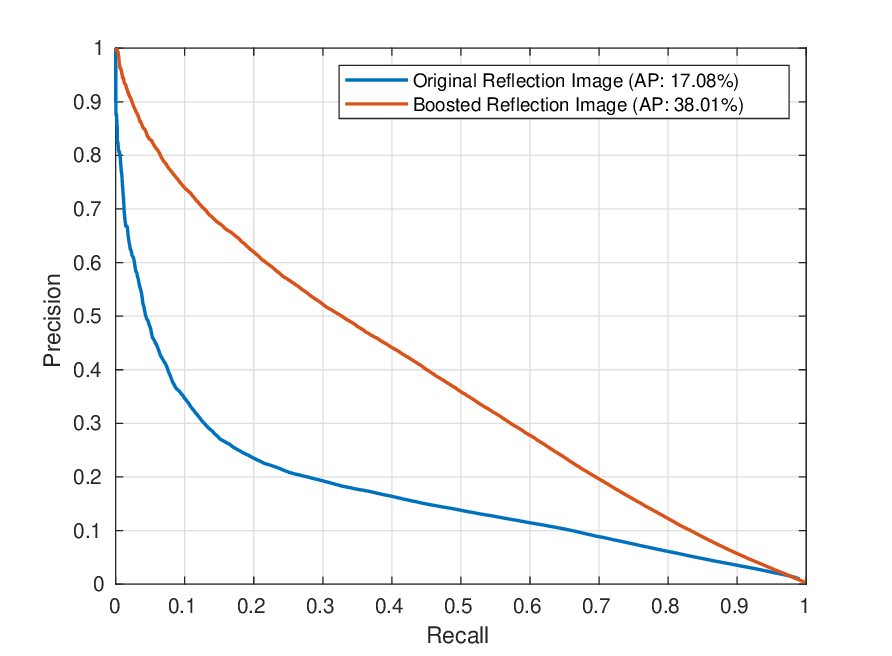}
   \caption{Reflection detection accuracy of the boosting DNN output image compared to the original intensity image.}
   \label{fig:boosting_dnn_accuracy}
\end{figure}

Fig.~\ref{fig:boosting_dnn_accuracy} presents the precision-recall curve of the boosting DNN output. The boosting DNN was trained as outlined in Section \ref{sec:res_object_det}. The precision-recall were calculated with a set of 5,000 synthetic examples. These examples were generated through the simulation detailed upon in Section \ref{sec:data_set} and were distinct from the training dataset. The figure offers a comparative perspective by presenting the precision-recall curve of the original radar reflection intensity images from the same test set, prior to undergoing the boosting DNN's influence.

The results presented in Fig.~\ref{fig:boosting_dnn_accuracy} clearly indicate a substantial enhancement in the accuracy of radar reflection detection achieved by the boosting DNN when compared to the original radar reflection intensity image. Notably, the average precision (AP) attained by the boosting DNN exceeds that of the original image by more than twofold. 
This improvement can be attributed to the boosting DNN's ability to generate a radar image that is both sharper and cleaner than the original image, as evidenced by the qualitative examples presented in Section \ref{sec:qualitative_perf} and in Section \ref{sec:qualitative_examples} below.

\subsection{Qualitative Evaluation With High-End Radar}\label{sec:qualitative_examples}
In Fig. \ref{fig:qualive_eval}, we showcase qualitative examples of \textit{BoostRad} using the RADDet dataset. This section presents additional qualitative examples, this time employing a higher-end radar compared to the one in RADDet. The specifications of this higher-end radar system can be found in Table ~\ref{tab:rad_spec}. Notably, it has three times higher range and four times higher azimuth angle resolution than the RADDet radar. We trained the boosting DNN using synthetic radar images produced via simulation of the higher-end radar. Subsequently, we tested both the boosting DNN and the object detection DNN using real measurements collected from this higher-end radar.

\begin{figure*}[!h]
  \centering
   \includegraphics[width=1\linewidth]{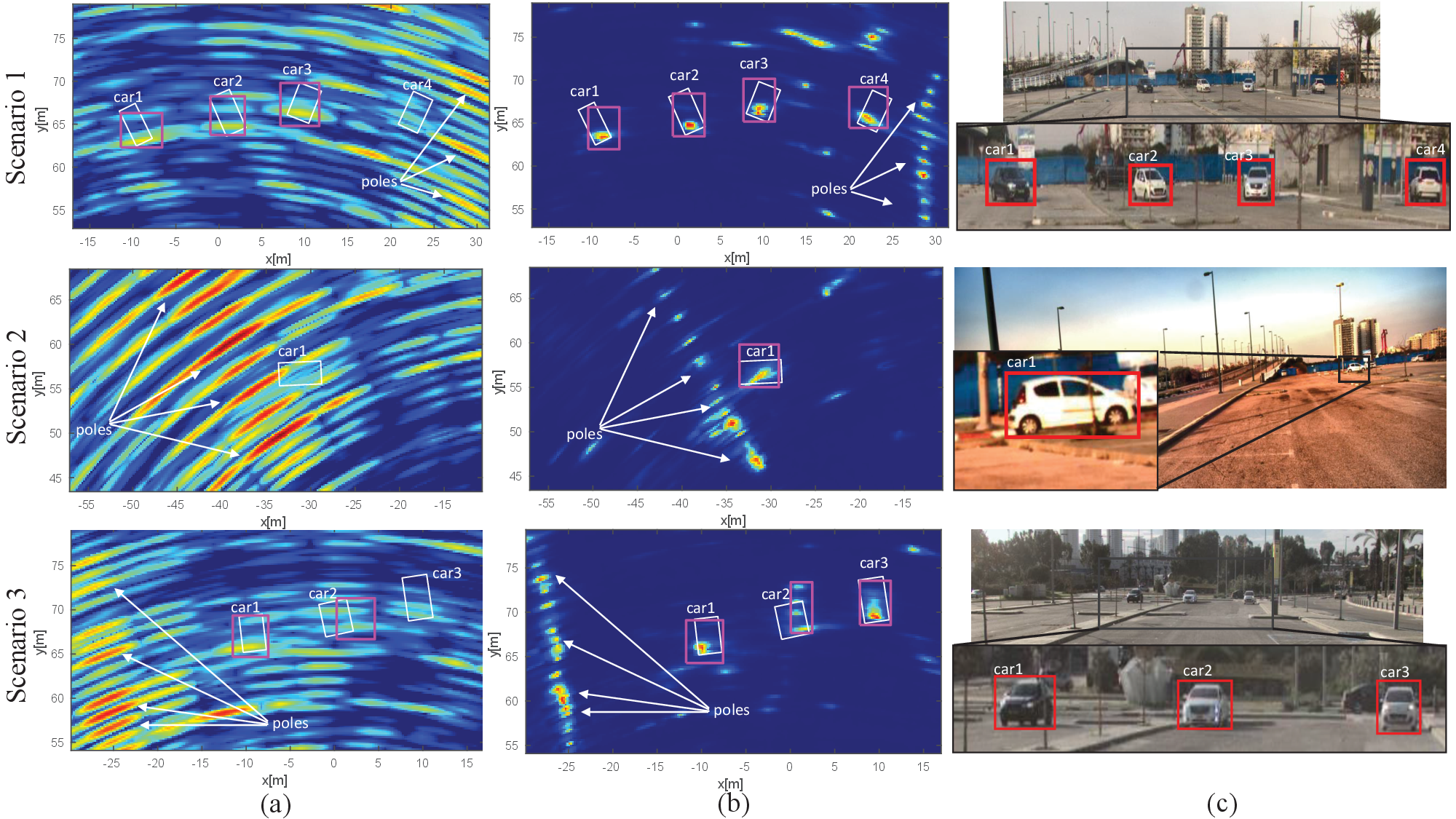}
   \caption{Qualitative examples with a higher-end radar than the radar used in RADDet. Each row is a different scenario. (a) original radar reflection image with the 'U-Net' reference detections, (b) boosting DNN output image with BoostRad detections, (c) camera image of the scene with ground truth bounding boxes. In (a) and (b), detection bounding boxes are purple, and ground truth bounding boxes are white. The vehicles have an identification number per each scenario that is written next to the ground truth bounding box.}
   \label{fig:qualitative_ex_m2r}
\end{figure*}

Fig.~\ref{fig:qualitative_ex_m2r} presents qualitative examples of real measurements from the higher-end radar. The original radar reflection image with the 'U-Net' object detection reference (see Section \ref{sec:res_object_det}) is presented in column (a). The output image of the boosting DNN and the BoostRad detected objects are presented in column (b), while column (c) showcases a scene photograph and an enlarged view of the scene's vehicles. Each row is a different scenario. In (a) and (b), the detected bounding boxes are denoted in purple, and the ground truth bounding boxes are denoted in white. In (c), the ground truth boxes are marked in red. Notably, the wide point spread function introduces clutter that obscures vehicles with lower reflectivity in (a). Consequently, this leads to the miss-detection of 'car 4' in Scenario 1, 'car 1' in Scenario 2, and 'car 3' in Scenario 3. In contrast, images in (b) exhibit considerably reduced clutter due to the narrower spreading function, facilitating the detection of all vehicles. Another observation of the narrower spreading function in (b) compared to (a) is apparent when examining the lighting poles that are located along a line on the right side of the image in scenario 1 and on the left side of the images in scenario 2 and 3 (as indicated in the figure). These lighting poles possess strong reflectivity, causing their spreading function main-lobe and side-lobes to stand out in (a). In contrast, (b) presents sharper lighting poles with significantly narrower spreading functions, resulting in reduced masking of other objects within the scene.

\begin{table}[h]
  \centering
  \begin{tabular}{@{}lc@{}}
    \toprule
    Radar parameter &  \\
    \midrule
    Maximal range & $150m$ \\    
    Range resolution & $0.28cm$ \\
    Azimuth field of view & $(-60^{\circ}:+60^{\circ})$\\
    Azimuth resolution & $3.9^{\circ}$ \\
    Elevation field of view & $(-20^{\circ}:+20^{\circ})$\\
    Elevation resolution & none\\
    Number of Tx antennas & 4 \\
    Number of Rx antennas & 8 \\
    Carrier frequency & $77GHz$ \\
    Sampling rate & $25Mhz$ \\
    Waveform & Fast chirps FMCW \\
    \bottomrule
  \end{tabular}
  \caption{Specifications of radar used in Sections \ref{sec:qualitative_examples}.}
  \label{tab:rad_spec}
\end{table}

\subsection{Object Detection AP at Relatively High IOU}\label{sec:ap_iou_large}
Table \ref{tab:results} presents Average Precision (AP) comparison between \textit{BoostRad} and the reference methods for IOU thresholds 0.1 and 0.3. Additional AP results are presented in Table \ref{tab:AP05} of the supplementary material, specifically at an IOU of 0.5 for the 'car' class within the RADDet dataset. As expected, the AP scores for all methods exhibit a reduction at an IOU of 0.5 in comparison to IOU values of 0.1 and 0.3. Nevertheless, the performance advantage of \textit{BoostRad} compared to the reference methods is still apparent. 
\begin{table}[]
\centering
\caption{Object Detection Average Precision on RADDet Dataset for class 'car' in all ranges ($R>0$) at IOU 0.5.}
\label{tab:AP05}
\begin{tabular}{ccc}
\toprule
Method                       & Boosting & @0.5   \\
\midrule
\multirow{3}{*}{RADDet \cite{zhang2021raddet}} & \xmark &47.98 \\
& \cmark & 50.70\\
&  & (\color{Green}{+2.72})\\
\midrule
\multirow{3}{*}{Probalistic \cite{dong2020probabilistic}} & \xmark &  40.68  \\
& \cmark & 45.16\\
& &  (\color{Green}{+4.48})\\
\midrule
\multirow{1}{*}{U-Net} & \xmark &   51.50 \\
\midrule
\multirow{2}{*}{BoostRad} &  \multirow{2}{*}{\cmark} & \textbf{52.75}\\
&  &  (\color{Green}{\textbf{+1.25}})\\
\midrule
\end{tabular}
\end{table}

\section{Radar Simulation Details}\label{sec:rad_sim_supp}
Section 3.3 of the primary paper offers a description of the simulation processing procedures. In this accompanying section, we expound upon the mathematical expressions employed in the simulation's implementation. The simulation involves a Multiple-Input Multiple-Output (MIMO) radar that emits a rapid chirp Frequency-Modulated Continuous-Wave (FMCW) waveform \cite{tong2015fast, bialer2021code}. The transmission signal as a function of time, $t$, is mathematically represented as follows:
\begin{equation}\label{TX_SIG}
x(t) = \sum_{m = 0}^{M-1} s(t-mT_c),
\end{equation}
where $T_c$ is the chirp duration, $M$ is the number of chirps per frame, and
\begin{equation}
s(t) =
    \begin{cases}
e^{-j2\pi (f_ct+\frac{1}{2}\alpha {t}^2)}, & \text{if}\ 0 \leq t \leq T_c \\
0, & \text{otherwise} \;,
\end{cases}
\end{equation}
is a single chirp, where $j=\sqrt{-1}$ is the imaginary unit, $f_c$ is the carrier frequency and $\alpha$ is the chirp slope.
The received signal at the $k^{th}$ receive antenna is a superposition of all the reflected signals from all transmit antennas, which is expressed by 
\begin{equation}\label{RX_SIG}
y_{k}(t) = \sum_{i = 0}^{I-1} \sum_{q = 0}^{Q-1} c^qx(t-\tau_{k,i}^q),
\end{equation}
where $Q$ are the number of reflection points, $c_q$ is the $q^{th}$ reflection point's intensity, $I$ is the number of transmit antennas, $i$ is the transmit antenna index, and $\tau_{k,i}^q$ is the round trip delay between the $i^{th}$ transmit antenna to the $q^{th}$ reflection point and back to the $k^{th}$ received antenna.

To derive the radar reflection intensity image, the received signal undergoes conventional FMCW radar processing \cite{bialer2021code}. This process encompasses down-conversion of the received signal through multiplication with the transmit reference signal \eqref{TX_SIG}, a range Fast Fourier Transform (FFT) for each chirp, a Doppler FFT across range bins over chirps, and beamforming for each range-Doppler bin across antennas.

\section{Additional Implementation Details}
Additional implementation details to those given in Section \ref{sec:res} are specified below. The probability mapping of the ground truth reference given  in \eqref{eq:p0} was calculated with a fixed noise variance $\sigma_n^2=8\times 10^{-5}$ and signal variance $\sigma_s^2=100R_{max}^{2}\sigma_n^2/r^2$, where $R_{max}=50[m]$ is the maximal range of the TI prototype radar (used in RADDet and CARRADA) and $r$ is the range of the ground truth reference pixel.
The hyper-parameters for the Boosting DNN loss in \eqref{eq:loss_boost} are: $(\rho_r,\rho_n,\rho_s)=(0.1,5,1)$. For the object detection loss, we weighted the L2 regression loss with factor $10^{-3}$ compared to the classification loss.

\section{RADDet Train-Test Set Split}
As outlined in Section \ref{sec:res}, the performance evaluation conducted in the main paper involved a train-test set split of the RADDet dataset that is different than the split proposed in \cite{zhang2021raddet}. Table \ref{tab:raddet_org_split} presents the results of \textit{BoostRad} and the reference methods on the original RADDet train-test split. The results show that all methods achieved higher results compared to Table \ref{tab:results}, and \textit{BoostRad} does not attain a performance gain compared to the reference methods. 
These results can be attributed to the fact that the test and training images in the original split \cite{zhang2021raddet} were derived from the same scenarios, often with small temporal gaps. As a result, a strong correlation is established between the test and training samples, leading to overfitting of all methods on the test set.

\begin{table*}[h]
\centering
\caption{Object Detection Average Precision on RADDet With Original Train-Test split in \cite{zhang2021raddet}}
\label{tab:raddet_org_split}
\begin{tabular}{ccccccccc}
\toprule
\multicolumn{1}{c}{} &       & \multicolumn{5}{c}{Car}                                                         & \multicolumn{2}{c}{Person}            \\
\cmidrule(lr){3-7} \cmidrule(lr){8-9}
\multicolumn{1}{c}{}                        &       & \multicolumn{3}{c}{$R \ge 0$} & \multicolumn{2}{c}{$R \ge 40$} & \multicolumn{2}{c}{$R \ge 0$} \\
\cmidrule(lr){3-5} \cmidrule(lr){6-7} \cmidrule(lr){8-9}
\multicolumn{1}{c}{Method}                        & Boosting & @ 0.1       & @ 0.3       & @ 0.5      & @ 0.1              & @ 0.3             & @ 0.1              & @ 0.3            \\
\midrule
RADDet \cite{zhang2021raddet} & \xmark & 93.82 & 88.03 & 68.71 & 85.76 & 62.73 & 69.40 &  43.48 \\
\midrule
Probalistic \cite{dong2020probabilistic} & \xmark & 92.87 & 86.36 & 66.30 &  81.55  &  58.58 & 64.29 & 48.10 \\
\midrule
\multirow{1}{*}{U-Net} & \xmark & 94.68 &  89.59 &   71.00 & 88.35 &  68.09  &  67.00 & 50.21 \\
\midrule
BoostRad & \cmark & 93.58 & 88.59 & 67.29 & 85.40 &  64.96  & 66.66 & 51.39 \\
\bottomrule
\end{tabular}
\end{table*}

To address the issue of overfitting, we adopted a train-test set partitioning approach that ensures that the training and testing sets encompass disparate scenarios. The RADDet dataset comprises 15 distinct scenes, each detailed in Table \ref{tab:RADDET_New_Split}. In our partitioning scheme, we designated scenes 9 and 11 for the test set, while the remaining scenes were utilized for the training set. Our modified training set encompasses a total of 17,021 cars (compared to the original partition's count of 16,755) and 5,240 pedestrians (compared to the original 5,210). For the test set, we have 4,094 cars (as opposed to 4,135 in the original) and 1,011 pedestrians (as opposed to 1,280).

\section{Probability Mapping Derivation}

\begin{table}[]
  \centering
  \begin{tabular}{cl}
    \toprule
    Scene ID & Frame Numbers  \\
    \midrule
0 & $0-439,559-724,1549-1971$\\
1 & $440-555,731-1548, 1972-2571$\\
2 & $2572-3038$ \\
3 & $3039-3437$\\
4 & $3438 -3653$\\
5 & $3654-4073$ \\
6 & $4074-4331$ \\
7 & $4332-5018, 5623 -6243$ \\
8 & $5019-5622,6244-6608$ \\
9 & $6609-8046$ \\
10 & $8047-8634$ \\
11 & $8635 - 9158$ \\
12 & $9159 -9437$ \\
13 & $9438-9745, 10175-10292$ \\
14 & $9746-10174$\\
    \bottomrule
  \end{tabular}
  \caption{RADDet partition into distinct scenes}
  \label{tab:RADDET_New_Split}
\end{table}

In this section we provide a detailed derivation of the mapping of ground truth 'super-radar' intensity to probability given in \eqref{eq:p0}. Denote by $z_i$ the complex value of the $i^{th}$ pixel of the ground truth reference image. We assume that $z_i$ is a complex Gaussian random variable with zero mean and variance that could be either of noise or a signal, denoted by $\sigma_n^2$ and $\sigma_s^2$, respectively. The intensity (energy) of a complex Gaussian random variable has Chi-square distribution with 2 degrees of freedom. Let $H_0$ and $H_1$ denote the hypotheses that the $i^{th}$ pixel is a noise or signal pixel, respectively. Then the probability density function of the $i^{th}$ pixels' intensity given each hypothesis can be expressed as
\begin{equation}\label{eq:sup_H0}
p(|z_i|^2\bigr|H_0) =  \frac{1}{2\sigma_{n}^2}e^{-|z_i|^2/(2\sigma_{n}^2)},
\end{equation}
\begin{equation}\label{eq:sup_H1}
p(|z_i|^2\bigr|H_1) =  \frac{1}{2\sigma_{s}^2}e^{-|z_i|^2/(2\sigma_{s}^2)}. 
\end{equation}

The probability mapping of the ground truth reference intensity is the posteriori probability density function $p(H_1| |z_i|^2)$. From Bayes rule and the law of total probability we have that 
\begin{multline}\label{eq:post_H1}
p(H_1\bigr||z_i|^2) = \frac{p(|z_i|^2\bigr|H_1)p(H_1)}{p(|z_i|^2)} = \\ \frac{p(|z_i|^2\bigr|H_1)p(H_1)}{p(|z_i|^2\bigr|H_1)p(H_1) + p(|z_i|^2\bigr|H_0)p(H_0)}.    
\end{multline}
By substituting \eqref{eq:sup_H0} and \eqref{eq:sup_H1} into \eqref{eq:post_H1}, and by assuming that $p(H_0)=p(H_1)=0.5$ we arrive at 
\begin{align}
     p(H_1\bigr||z_i|^2) = \frac{e^{-|z_i|^2/(2\sigma_{s_i}^2)}}{e^{-|z_i|^2/(2\sigma_s^2)}+\frac{\sigma_{s_i}^2}{\sigma_n^2}e^{-|z_i|^2/(2\sigma_n^2)}},
\end{align}
which is the probability mapping given in \eqref{eq:p0}.

\end{document}